# How People Respond to the COVID-19 Pandemic on Twitter: A Comparative Analysis of Emotional Expressions from US and India

Brandon Siyuan Loh[1], Raj Kumar Gupta[1], Ajay Vishwanath[1,2], Andrew Ortony[3], Yinping Yang[1]*

[1]Institute of High Performance Computing (IHPC), Agency for Science, Technology and Research (A*STAR), Singapore

[2]Centre for Artificial Intelligence Research, University of Agder, Norway

[3]Department of Psychology, Northwestern University, Evanston, Illinois, United States

*Correspondence: Yang Yinping (yangyp@ihpc.a-star.edu.sg)

## Abstract

The COVID-19 pandemic has claimed millions of lives worldwide and elicited heightened emotions. This study examines the expression of various emotions pertaining to COVID-19 in the United States and India as manifested in over 54 million tweets, covering the fifteen-month period from February 2020 through April 2021, a period which includes the beginnings of the huge and disastrous increase in COVID-19 cases that started to ravage India in March 2021. Employing pre-trained emotion analysis and topic modeling algorithms, four distinct types of emotions (fear, anger, happiness, and sadness) and their time- and location-associated variations were examined. Results revealed significant country differences and temporal changes in the relative proportions of fear, anger, and happiness, with fear declining and anger and happiness fluctuating in 2020 until new situations over the first four months of 2021 reversed the trends. Detected differences are discussed briefly in terms of the latent topics revealed and through the lens of appraisal theories of emotions, and the implications of the findings are discussed.

## Keywords

COVID-19, pandemic, emotion, social media data, appraisal theory, fear, anger, happiness, sadness, Twitter

## Introduction

According to the World Health Organization's situation dashboard[1] (WHO, 2021), as of 6 May 2021, the COVID-19 pandemic had infected 155,665,214 people worldwide and claimed 3,250,648 lives. At that time, the United States and India were the first and second worst-hit countries, with 32,210,817 confirmed cases and 573,722 deaths in the US, and in India, 21,491,598 reported confirmed cases, and 234,083 deaths. The pandemic severely threatens people's physical and emotional well-being (Banerjee and Rai, 2020; Brooks et al., 2020; Saladino et al., 2020; van Bavel et al., 2020).

In the work we describe, we sought preliminary insight into people's expressed emotional reactions to the pandemic. This issue is important for at least two general reasons. First, different emotions have distinct effects on risk perception and risk-related behavior. For instance, Lerner and Keltner (2001) observed that angry people harbored optimistic risk estimates, driving them toward risky choices, while fearful people expressed pessimistic risk perceptions and were more inclined toward risk-averse behavior. Second, there are social and interpersonal consequences of emotional expressions. The tendency of people to 'take on' the emotional states of others (i.e., emotional contagion; Hatfield et al., 1993; Kramer et al., 2014), and to rely on the emotional expressions of others when forming judgments or making decisions (van Kleef, 2009) is especially relevant as it highlights the susceptibility to external social influences of people's judgments and reactions.

Several studies have attempted to describe how people have responded to the pandemic. Abd-Alrazaq et al. (2020) described the top concerns of Tweeters. Budhwani and Sun (2020) examined the use of pejorative phrases such as "Chinese virus". Lwin et al. (2020) provided a rapid report on global emotion trends, showing that anger overtook fear during the very early phase of the pandemic (February to April 2020), and Garcia and Berton (2021) analyzed topics and sentiments in tweets from the US and Brazil (April to August 2020). However, to our knowledge, no studies have contrasted temporal changes in pandemic-relevant emotion expressions across different countries, and none has attempted to explain such dynamics in psychological terms.

## Methods

We obtained the raw data from a publicly available labeled tweets database[2] gathered via calls to Twitter's standard search API. Only English tweets containing at least one COVID-related keyword or word stem, such as 'covid' and 'corona' were included. Retweets were excluded. This database is described in detail by Gupta et al. (2021).

For this study, we selected tweets whose user profiles suggested that they (the users) were either from the United States or India. We used a city-to-country database (GeoNames, 2000) to match each tweet author's self-reported location to a corresponding country code. For example, tweets with locations disclosed as "Washington, DC" were classified as "United States", while "Tamil Nadu, India" was parsed as "India".

We applied CrystalFeel[3], a set of linear SVM-based emotion analysis algorithms trained to infer the intensity of four distinct types of emotions, namely, fear, anger, happiness and sadness, from each tweet (Gupta and Yang, 2018). CrystalFeel uses both affective features (extracted using various sentiment and emotion lexicons) and non-affective linguistic features (e.g., parts-of-speech, pre-trained word embeddings) to predict the intensity of each emotion on a *continuous* scale from 0 (not at all) to 1 (extremely intense). Evaluated on the SemEval-18 affect in tweets shared task data (Mohammad et al., 2018), the algorithms' emotion intensity outputs reported consistently high Pearson correlations with human labels: 0.70 (fear), 0.74 (anger), 0.71 (happiness), and 0.72 (sadness) (Gupta and Yang, 2018). Based on the intensity scores, CrystalFeel generates emotion classification outputs, which correspond to predicting a 5-class discrete emotion type ("fear", "anger", "happiness", "sadness", and "no specific emotion or neutral") that the tweet expresses. (Appendix A presents details about the methods of emotion extraction and examples of tweets with their corresponding emotion outputs.)

---

[1] https://covid19.who.int
[2] https://doi.org/10.3886/E120321
[3] https://socialanalyticsplus.net/crystalfeel

In addition to emotional information, we extracted information about the topics covered by the tweets, expecting that we would eventually be able to analyze in more detail what emotions were being expressed and with what intensity and what the emotions were about. To this end, we used jLDADMM[4] (Nguyen, 2018) to infer topics present in the tweets. The jLDADMM package is based on Dirichlet Multinomial Mixture (DMM), a probabilistic generative model designed for short text clustering (Nigam et al., 2000). Unlike traditional topic models, where each word in a document is assumed to be generated from a distribution of topics (Blei et al., 2003), DMM takes a generative process such that there is a one-to-one correspondence between the mixture components and topics (Yin and Wang, 2014). By limiting each document to one topic, DMM overcomes the challenge of data sparsity and limited contextual information in modeling short texts (Nguyen et al., 2015). (Appendix B reports the details of the extracted topics.)

The processed data comprises 54,941,724 tweets over fifteen months (28 January 2020 through 1 May 2021): 47,037,387 tweets from the United States and 7,904,337 tweets from India-based users. Figures 1 and 2 depict the volume of the daily aggregated tweets and daily proportions of emotions for the United States and India, respectively. We have highlighted six-time points in Figures 1 and 2 that are particularly interesting because of salient changes in the distribution of emotions expressed around those times. The data and dashboard used for this study are available at http://52.3.21.155/covid2019/in_us.php.

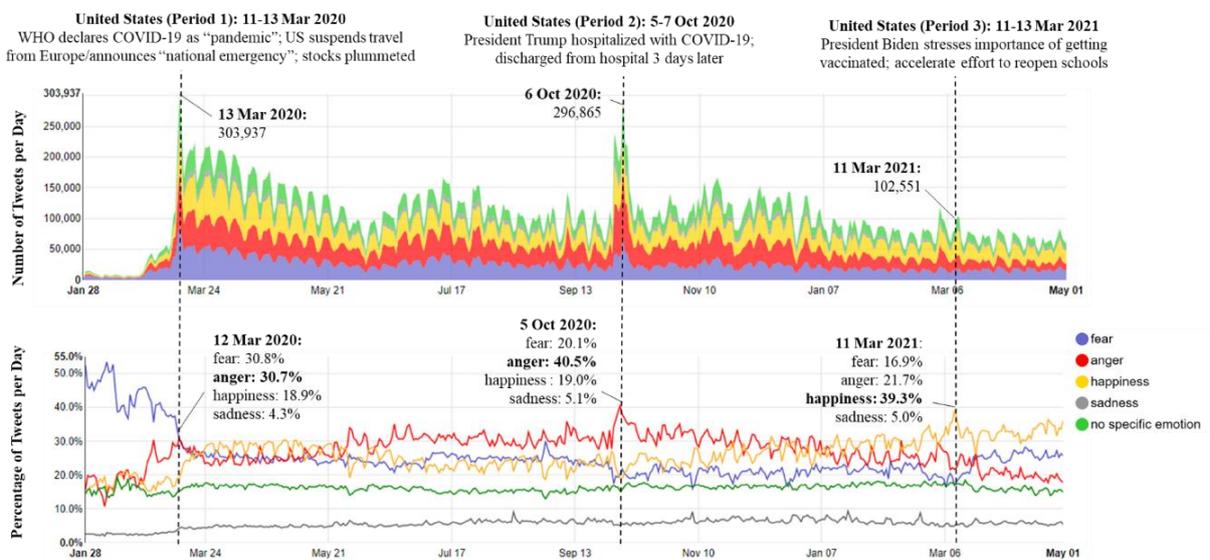

**Figure 1**. United States—Daily counts of tweets at five levels of sentiment valence (upper panel), and daily percentage of tweets across four types of emotions (lower panel)

---

[4] https://github.com/datquocnguyen/jLDADMM

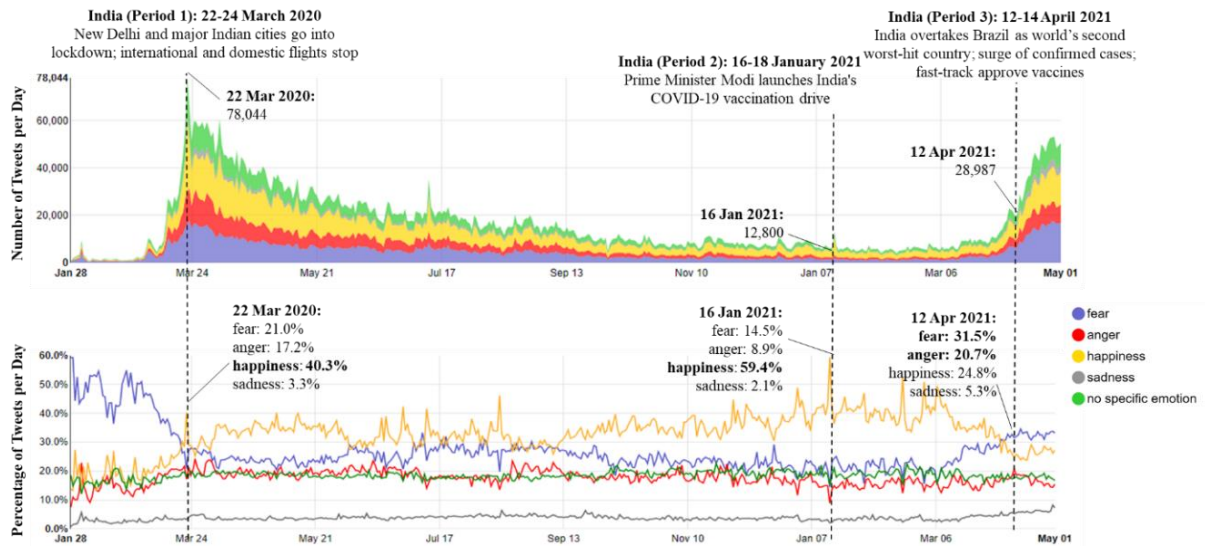

**Figure 2**. India—Daily counts of tweets at five levels of sentiment valence (upper panel), and daily percentage of tweets across four types of emotions (lower panel)

Next, we used segmented regression analysis to determine whether the detected differences were statistically significant. We set each date as the point of 'intervention'. We conducted a separate segmented regression analysis to estimate whether there were differences in the level and trend of each emotion-type seven days pre- and post- 'intervention'.

Finally, we performed visual analysis ("emotion-topic heat map") to analyze the patterns of emotion distribution per each topic, as well as the patterns of topic distribution per each emotion.

**Results**

A notable pattern that surfaced from our data concerns variations in the relative proportions of fear, anger, happiness, and sadness emotions following the first wave of the pandemic (i.e., March 2020). The upper panel of Figure 1 shows that the number of tweets posted from the US was at its all-time high on 13 March 2020 ($n$ = 303,937 tweets). At the same time, there was an escalation in the percentage of tweets expressing *anger* (30.4%; see Figure 1, lower panel), and 13 March 2020 seems to be the date where angry tweets started to replace fearful tweets (30.7%). On 6 October 2020, *anger* peaked daily percentage of 40.1% of 296,865 tweets posted that day. Fast forward to 11 March 2021, *happiness* reached its peak value of 39.3%. In contrast, in India, from 22 March 2020, when the number of tweets reached the daily peak value of 78,044 (Figure 2, upper panel), the percentage of *happiness* increased consistently throughout the year and into the first two months of 2021 (Figure 2, lower panel), with the portion of *happiness* reaching the country's historical peak of 59.4% on 16 January. However, *happiness* in India showed a dramatic downward trend starting in late March 2021, and on 12 April had fallen to 24.8%, while *fear*, at 31.5%, began to rise again above the 30% mark.

Results of segmented regression analysis suggested that, except for *fear* in all three periods and *sadness* in Period 3, the after-event differences of *anger*, *happiness*, and *sadness* were statistically significant for the US. For India, the only significant after-event difference is *fear* in Period 1. Table 1 presents the statistical significance analysis of the time-specific trends.

**Table 1**. Regression results using the aggregated percentage of tweets for each emotion as the criterion variable and time as the predictor variable

|  |  | US (Period 1): 12 March 2020 | | | US (Period 2): 6 October 2020 | | | US (Period 3): 11 March 2020 | | |
|---|---|---|---|---|---|---|---|---|---|---|
|  |  | b | t | p-value | b | t | p-value | b | t | p-value |
| fear | Time | -0.010 | -4.354 | .001** | -0.007 | -2.618 | .024* | 0.001 | 0.526 | .609 |
|  | Event | -0.036 | -2.876 | .015* | 0.022 | 1.567 | .145 | -0.022 | -2.152 | .054 |
|  | Time_after_event | 0.005 | 1.666 | .123 | 0.005 | 1.416 | .185 | 0.004 | 1.762 | .106 |
| anger | Time | 0.010 | 4.822 | .001** | 0.021 | 6.271 | .000** | -0.012 | -4.445 | .001** |
|  | Event | -0.015 | -1.375 | .197 | -0.023 | -1.254 | .236 | 0.006 | 0.380 | .712 |
|  | **Time_after_event** | **-0.013** | **-4.845** | **.001\*\*** | **-0.026** | **-6.162** | **.000\*\*** | **0.015** | **4.202** | **.001\*\*** |
| happiness | Time | -0.002 | -0.738 | .476 | -0.011 | -3.660 | .004** | 0.010 | 2.743 | .019* |
|  | Event | 0.036 | 3.065 | .011* | -0.010 | -0.621 | .547 | 0.018 | 0.982 | .366 |
|  | **Time_after_event** | **0.008** | **2.875** | **.015\*** | **0.017** | **4.655** | **.001\*\*** | **-0.017** | **-3.870** | **.003\*\*** |
| sadness | Time | 0.002 | 4.415 | .001** | -0.002 | -3.271 | .007** | 0.000 | 0.602 | .560 |
|  | Event | 0.006 | 2.464 | .031* | 0.005 | 1.312 | .216 | 0.002 | 0.507 | .622 |
|  | **Time_after_event** | **-0.002** | **-4.323** | **.001\*\*** | **0.002** | **2.700** | **.021\*** | **-0.001** | **-0.920** | **.377** |
| no specific emotion | Time | 0.000 | -0.067 | .820 | -0.001 | -0.691 | .504 | 0.001 | 1.401 | .189 |
|  | Event | 0.010 | 1.668 | .118 | 0.005 | 0.720 | .486 | -0.004 | -0.738 | .476 |
|  | Time_after_event | 0.002 | 1.654 | .198 | 0.002 | 0.959 | .358 | 0.001 | -0.744 | .472 |
|  |  | India (Period 1): 22 March 2020 | | | India (Period 2): 16 January 2021 | | | India (Period 3): 12 April 2021 | | |
|  |  | b | t | p-value | b | t | p-value | b | t | p-value |
| fear | Time | 0.054 | 2.287 | .048 | -0.006 | -0.380 | .713 | 0.002 | -0.304 | .768 |
|  | Event | 0.093 | 1.932 | .085 | -0.071 | -2.281 | .049* | 0.025 | -2.312 | .046* |
|  | **Time_after_event** | **-0.084** | **-2.449** | **.039\*** | **0.038** | **1.701** | **.123** | **0.013** | **1.719** | **.120** |
| anger | Time | -0.013 | -1.140 | .283 | -0.001 | -1.252 | .912 | 0.003 | 0.641 | .538 |
|  | Event | -0.001 | -0.041 | .968 | -0.068 | -4.658 | .001** | 0.025 | 2.978 | .016* |
|  | Time_after_event | 0.011 | 0.675 | .517 | 0.019 | 1.803 | .105 | -0.008 | -1.287 | .230 |
| happiness | Time | -0.009 | -0.348 | .736 | 0.019 | 0.743 | .477 | -0.004 | -0.309 | .764 |
|  | Event | -0.064 | -1.265 | .238 | 0.192 | 3.653 | .005** | 0.002 | 0.091 | .930 |
|  | Time_after_event | 0.026 | 0.717 | .491 | -0.091 | -2.425 | .038 | -0.004 | -0.199 | .847 |
| sadness | Time | 0.000 | -0.148 | .886 | 0.002 | 0.108 | .916 | 0.003 | 0.996 | .345 |
|  | Event | 0.011 | 2.304 | .047* | -0.014 | -3.097 | .0013* | -0.003 | -0.558 | .590 |
|  | Time_after_event | -0.001 | -0.401 | .697 | 0.004 | 1.231 | .250 | -0.001 | -0.314 | .760 |
| no specific emotion | Time | 0.003 | -3.444 | .007** | 0.015 | -1.440 | .183 | -0.001 | -0.211 | .838 |
|  | Event | -0.040 | -2.058 | .070 | -0.038 | -1.848 | .098 | -0.002 | 0.163 | .874 |
|  | **Time_after_event** | **0.049** | **3.580** | **.001\*\*** | **0.030** | **2.065** | **.069** | **0.001** | **0.066** | **.949** |

Note: The "Time" coefficient estimates the trend before the event. "Event" is a dummy variable coding for whether a particular time point occurs before or after a critical event, with its coefficient assessing the post-event intercept. "Time_after_event" represents the number of time steps after the critical event, with its coefficient representing the change in trend over the pre- and post-critical event slopes. *b* represents unstandardized regression weights. *t* is the coefficient divided by its standard error. For *p-value*, * indicates $p < .05$. ** indicates $p < .01$.

Finally, results combining topics and emotions in the form of a heat map (Figure 3) provide further insights into the emotions of interest at particular time points in both countries.

For example, and notably, for the peak of *anger* in the US (Period 2), Topic 9 (trump, presid, biden) accounts for 20.0%, and Topic 1 (trump, hous, white) accounts for 13.1%, of the total angry tweets. For the sign of rising *fear* detected in India (Period 3), Topic 5 (vaccin, india, hashtagcovid), accounts for 19.0%, and Topic 4 (rally, elect, peopl) accounts for 15.7%. These topic-emotion correspondences provide insight into which emotions emerge and how they evolve.

**US (Period 1): 11-13 March 2020**

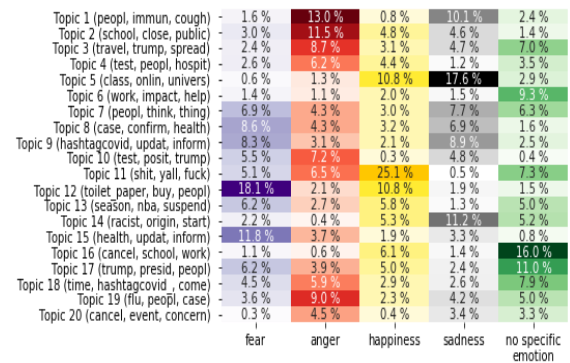

**India (Period 1): 22-24 March 2020**

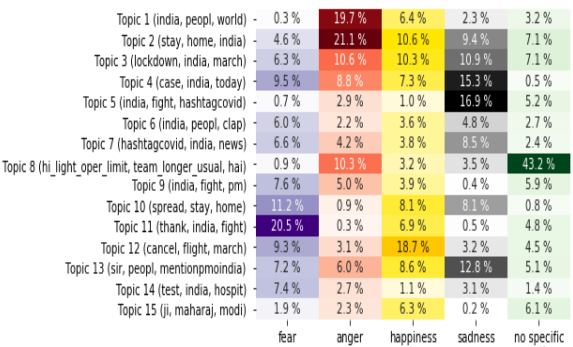

**US (Period 2): 6-8 October 2020**

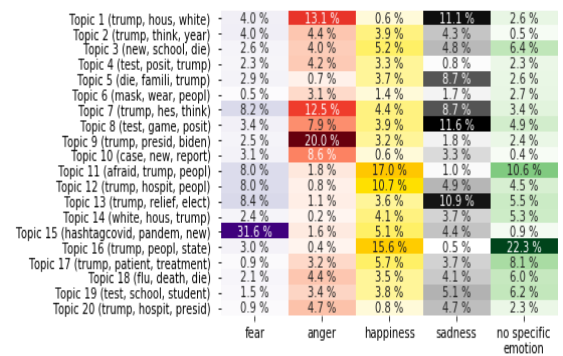

**India (Period 2): 16-18 January 2021**

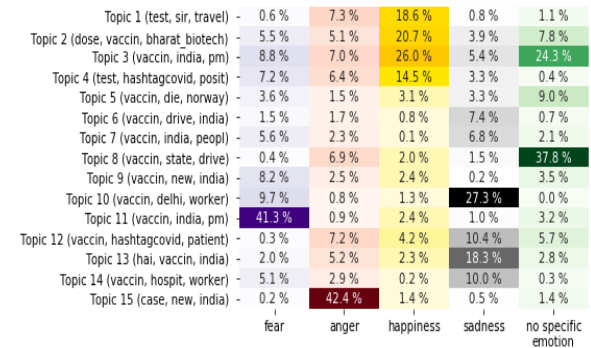

**US (Period 3): 11-13 March 2021**

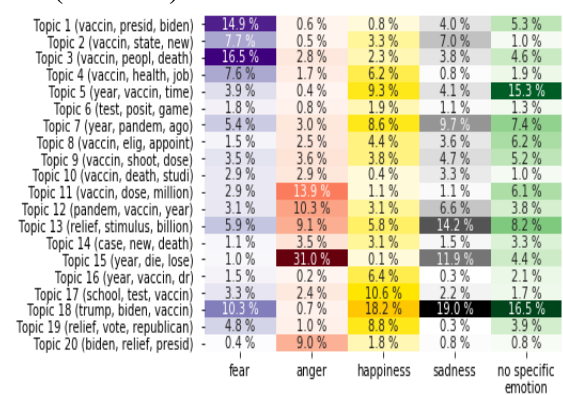

**India (Period 3): 12-14 April 2021**

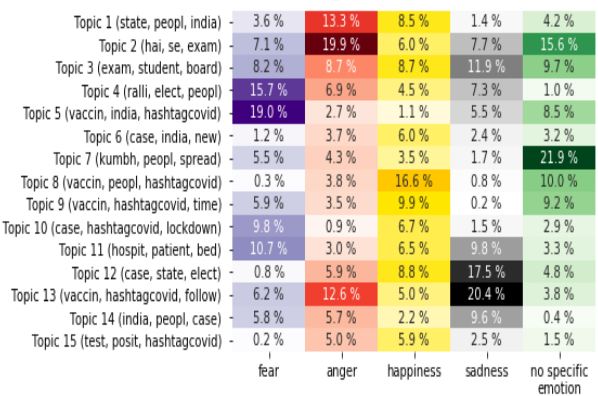

Note: the words or word stems alongside each topic represent the three most frequently associated with each detected topic

**Figure 3**.  The emotion-topic heat maps

## Discussion

Our general theoretical orientation is that of appraisal theories (Lazarus, 1991; Ortony et al., 1988; Roseman, 1996; Scherer et al., 2001; Smith and Ellsworth, 1985), the central idea of which is that different types of emotions arise as a result of different cognitive evaluations (appraisals) of the situations that elicit them. Consistent with such theories is the fact that as new information surfaces, the cognitions involved in appraising a protracted event often change, leading not only to changes in emotion intensities but also to the emergence of different emotions. Results of the current investigation demonstrating shifts in the relative proportions and intensities of the four emotions as the pandemic unfolded attest to this fact.

Apart from cognitive considerations, internalized culture-specific beliefs may also influence the way emotions are experienced and expressed. Although the practical situation for people in the US and India during the pandemic was in many ways similar (e.g., sudden loss of loved ones, lockdowns), some of the emotion-related differences between the countries might well be due to cultural differences in self-construal, that is, to culturally influenced beliefs about the relation of the self to others (Cross et al., 2011; Markus and Kitayama, 1991). For example, people from the US might be more sensitive than those in India to perceived government-imposed threats to their sense of individual freedom and autonomy. Hence, they might experience and report more anger because, unlike many of their Indian counterparts, they evaluate such restrictions as impediments to the pursuit of their personal goals. Future research might explore such ideas.

As well as being socially contagious, emotions influence all manner of behaviors, not least of which is risk-taking behaviors. For example, it is clearly the case that in the US, emotions have played a role in many people defying mask-wearing and social distancing mandates. The work we have described shows how applying advanced emotion analysis and automatic topic detection algorithms can illuminate important macro-level aspects of people's emotional experiences as they unfold during the pandemic. We hope this can pave the way for a better understanding of which particular emotion types have what social and behavioral consequences, where, and under what conditions during a prolonged, large-scale event.

## Acknowledgments

This work was supported by the Agency for Science, Technology and Research (A*STAR), Singapore, under its A*ccelerate Gap Fund (ETPL/18-GAP050-R20A) and A*CRUSE (COVID-19 social media analytics project).

## Author Contributions

SL initiated the study and performed data analysis. RG downloaded the full dataset and extracted the country and emotion features. AV performed the topic modeling and analysis. AO contributed to the interpretation and discussion of the results. YY conceptualized and designed the analytic framework. All authors contributed to the manuscript writing, reviewed the content, and agreed with the submission.

# Appendix A    Details of Emotion Extraction Methods and Examples

According to CrystalFeel's analytic scheme, each emotion label covers a family of related feelings. The emotion intensity score ranges from 0 to 1, where 0 = the text does not express this emotion at all; 1 = the text expresses an extremely high intensity of this emotion. (Source: https://socialanalyticsplus.net/crystalfeel)

- "anger" covers a family of anger-related negative feelings such as annoyance, irritation, aggravation, resentment, reproach, disliking, anger, fury, and rage

- "fear" covers a family of fear-related negative feelings such as apprehension, anxiety, worry, scared, dread, horror, and terror

- "sadness" covers a family of sadness-related negative feelings such as distress, helplessness, disappointment, melancholy, sorry-for, self-approach, remorse, sorrow, and grief

- "joy" or "happiness" covers a broad family of positive emotions such as joy, happy-for(-others), satisfaction, relief, appreciation, gratification, liking, contentment, pleasure, happiness, ecstasy, and excitement, as well as some subtle sense of hope, pride, gratitude, and compassion

- In addition, the "valence" output dimension covers a range of overall feelings from 0 = extremely unpleasant/negative to 1 = extremely pleasant/positive.

Table A1 presents the default setting of the conversion logic from CrystalFeel used for this study. Table A2 shows examples of tweets and the corresponding emotion intensity scores and emotion/sentiment types outputs.

**Table A1.** The conversion logic that generates 5-class emotion outputs

```
1.  # Initialize the emotion category in a no specific emotion class
2.         emotion = "no specific emotion";
3.  # Assign the emotion category when valence intensity score exceeds 0.52
4.  if (valence_intensity > 0.52):
5.         emotion = "happiness";
6.  # Assign the emotion category when valence intensity score falls below 0.48
7.  elif (valence_intensity < 0.48):
8.         emotion = "anger";
9.     if ((fear_intensity > anger_intensity) and (fear_intensity > = sadness_intensity )):
10.           emotion = "fear";
11.    elif ((sadness_intensity > anger_intensity) and (sadness_intensity > fear_intensity)):
12.           emotion = "sadness";
```

**Table A2.** Example emotion analysis outputs

| | Tweet (example) | country | fear_intensity | anger_intensity | sad_intensity | happiness_intensity | valence_intensity | emotion |
|---|---|---|---|---|---|---|---|---|
| 1 | This Covid-19 dilemma got my anxiety on a thousand and me scared to go outside or be near people. | United States | **0.931** | 0.520 | 0.682 | 0.168 | 0.229 | **fear** |
| 2 | This is trash!!! F**k you corona! | United States | 0.504 | **0.834** | 0.602 | 0.228 | 0.233 | **anger** |
| 3 | Corona is giving me a whole new level of anxiety and fear. | India | **0.989** | 0.550 | 0.673 | 0.186 | 0.191 | **fear** |
| 4 | @IndiaToday He deserves slapping for attempting to run the public morale down He is more dangerous than the corona itself | India | 0.550 | **0.598** | 0.496 | 0.139 | 0.293 | **anger** |
| 5 | Mother of global crisis.. who will visit a corona City | India | **0.541** | 0.435 | 0.462 | 0.236 | 0.377 | **fear** |
| 6 | Why are people arguing over coronavirus? Even if the flu also kills people, the hospitals in NY are over run right | United States | **0.591** | 0.562 | 0.499 | 0.159 | 0.287 | **fear** |
| 7 | He is Back..and this time he has become a cult hero ....Corona Go :))) | India | 0.234 | 0.295 | 0.316 | **0.578** | 0.684 | **joy** |
| 8 | WFH has some perks too. #chlear #marketing #advertising #agencylife #creative #advertisingagency #bangaloreagency | India | 0.220 | 0.220 | 0.247 | **0.423** | 0.695 | **joy** |
| 9 | Saddened by first Reported #Covid Death in #Goa; condolences to the family. https://t.co/TqepPokmg6 | India | 0.627 | 0.475 | **0.716** | 0.191 | 0.307 | **sad** |
| 10 | Is it possible that we all can be positive and helpful in this health crisis? #Covid_19 We all can use a little help in that department. | United States | 0.410 | 0.298 | 0.377 | 0.326 | 0.490 | **no specific emotion** |

# Appendix B  Details on Topic Analysis Results

Tables B1 and B2 present automatic topic modeling results for the three eriods selected for the United States and India-based tweets, respectively. We used the default parameters in the jLDADMM library for topic modeling, alpha = 0.1, beta - 0.01, niters = 2000. As the tweets from India on those specific days resulted in a smaller dataset than tweets from USA, we opted to extract 15 topics compared to the default value of 20. Pre-processing steps—including removing urls, non ascii words and stemming—follow the steps in the COVID-19 topic modeling presented by Gupta et al. (2021).

**Table B1**. United States—Automatic topic modeling results for three selected periods

| | | | | | | | | | | |
|---|---|---|---|---|---|---|---|---|---|---|
| | US (Period 1): 11-13 March 2020 | | | | | | | | | |
| Topic 1 | peopl | think | shit | yall | die | thing | fuck | flu | joke | kill |
| Topic 2 | toilet_paper | peopl | hashtagcovid_ | buy | need | drink | think | yall | eat | immun |
| Topic 3 | updat | emerg | declar | live | trump | presid | state | nation | respons | news |
| Topic 4 | cancel | event | postpon | concern | updat | march | announc | suspend | schedul | sport |
| Topic 5 | travel | trump | spread | pandem | hashtagcovid_19 | ban | countri | peopl | state | hashtagcovid |
| Topic 6 | health | spread | pandem | safeti | public | communiti | monitor | concern | close | organ |
| Topic 7 | trump | hashtagcovid | presid | peopl | test | respons | think | democrat | need | tri |
| Topic 8 | test | kit | free | hashtagcovid_ | need | health | peopl | trump | cdc | state |
| Topic 9 | hashtagcovid_19 | hashtagcovid | time | spread | help | peopl | need | stay | read | good |
| Topic 10 | test | symptom | peopl | patient | hospit | risk | doctor | flu | sick | infect |
| Topic 11 | flu | peopl | case | death | die | infect | number | kill | million | year |
| Topic 12 | work | cancel | school | home | email | week | close | peopl | time | hashtagcovid_ |
| Topic 13 | updat | inform | resourc | hashtagcovid | respons | help | latest | school | visit | websit |
| Topic 14 | school | close | march | class | week | onlin | cancel | student | univers | updat |
| Topic 15 | shit | fuck | cancel | yall | aint | gonna | catch | come | damn | bitch |
| Topic 16 | season | nba | cancel | suspend | game | test | play | player | posit | hashtagcovid_19 |
| Topic 17 | test | posit | trump | tom_hank | wife | hashtagcovid | peopl | presid | state | meet |
| Topic 18 | case | confirm | test | state | counti | health | new | posit | report | updat |
| Topic 19 | hashtagcovid_19 | time | come | think | new | gonna | watch | year | good | need |
| Topic 20 | racist | origin | start | media | come | news | peopl | trump | think | flu |
| | US (Period 2): 6-8 October 2020 | | | | | | | | | |
| Topic 1 | trump | hous | white | debat | infect | peopl | penc | test | wh | spread |
| Topic 2 | trump | think | year | peopl | time | feel | hes | look | come | man |
| Topic 3 | new | school | die | close | order | state | nyc | arkansa | busi | case |
| Topic 4 | test | posit | trump | negat | stephen_miller | result | presid | quarantin | didnt | debat |
| Topic 5 | die | famili | trump | home | think | peopl | hospit | lose | presid | nurs |
| Topic 6 | mask | wear | peopl | spread | trump | cdc | work | protect | mentionjoebiden | think |
| Topic 7 | trump | hes | think | fake | presid | peopl | lie | doesnt | believ | biden |
| Topic 8 | test | game | posit | nfl | team | player | season | play | week | patriot |
| Topic 9 | trump | presid | biden | diagnosi | debat | campaign | joe | poll | news | hospit |
| Topic 10 | case | new | report | death | state | counti | posit | updat | number | health |
| Topic 11 | afraid | trump | peopl | die | american | let | tell | live | dead | famili |
| Topic 12 | trump | hospit | peopl | patient | hes | learn | secret_servic | presid | infect | drive |
| Topic 13 | trump | relief | elect | talk | stimulus | negoti | senat | american | vote | presid |
| Topic 14 | white | hous | trump | test | posit | press_secreta | presid | report | return | new |
| Topic 15 | hashtagcovid | pandem | new | vaccin | health | work | help | impact | need | time |
| Topic 16 | trump | peopl | state | death | presid | american | vote | think | democrat | respons |
| Topic 17 | trump | patient | treatment | steroid | doctor | hes | drug | symptom | hospit | dexamethason |
| Topic 18 | flu | death | die | peopl | trump | year | kill | cdc | number | dead |
| Topic 19 | test | school | student | health | week | new | today | state | case | posit |
| Topic 20 | trump | hospit | presid | afraid | medic | treatment | care | leav | walter_reed | best |
| | US (Period 3): 11-13 March 2021 | | | | | | | | | |
| Topic 1 | vaccin | presid | biden | american | new | trump | relief | live | urg | thank |
| Topic 2 | vaccin | state | new | restrict | nurs_home | governor | texa | cdc | peopl | biden |
| Topic 3 | vaccin | peopl | death | think | trump | state | die | biden | year | mask |
| Topic 4 | vaccin | health | job | join | test | updat | march | communiti | hashtagcovid | help |
| Topic 5 | year | vaccin | time | think | peopl | thing | work | today | come | test |
| Topic 6 | test | posit | game | team | season | duke | year | play | tournament | cancel |
| Topic 7 | year | pandem | ago | today | anniversari | mark | declar | world | march | chang |
| Topic 8 | vaccin | elig | appoint | state | counti | health | peopl | receiv | dose | new |
| Topic 9 | vaccin | shoot | dose | second | today | feel | receiv | nd | effect | moderna |
| Topic 10 | vaccin | death | studi | effect | state | rat | new | peopl | risk | hashtagcovid |
| Topic 11 | vaccin | dose | million | biden | johnson_johnson | administ | astrazeneca | state | jandj | blood_clot |
| Topic 12 | pandem | vaccin | year | hashtagcovid | health | new | work | help | need | impact |
| Topic 13 | relief | stimulus | billion | american | new | trillion | check | packag | fund | pass |
| Topic 14 | case | new | death | report | counti | updat | health | march | state | total |
| Topic 15 | year | die | lose | famili | peopl | vaccin | hospit | test | friend | month |
| Topic 16 | year | vaccin | dr | join | new | pandem | hashtagcovid | today | discuss | work |
| Topic 17 | school | test | vaccin | year | student | counti | open | case | new | reopen |
| Topic 18 | trump | biden | vaccin | presid | american | peopl | death | thank | joe | credit |
| Topic 19 | relief | vote | republican | biden | american | democrat | pass | trillion | senat | support |
| Topic 20 | biden | relief | presid | trillion | sign | hous | pass | stimulus | packag | joe |

**Table B2.** India—Automatic topic modeling results for three selected periods

| | | | | | | | | | |
|---|---|---|---|---|---|---|---|---|---|
| India (Period 1): 22-24 March 2020 | | | | | | | | | |
| Topic 1 | pandem | declar | hashtagcovid | world | hashtagcorona | india | spread | health | updat | time |
| Topic 2 | case | india | death | hashtagcovid | countri | confirm | number | total | updat | new |
| Topic 3 | ipl | postpon | india | cancel | hashtagipl | match | suspend | hashtagcorona | hashtagcovid | till_april |
| Topic 4 | india | death | case | confirm | karnataka | die | posit | hashtagcovid | report | delhi |
| Topic 5 | india | world | fight | govt | countri | peopl | indian | spread | modi | govern |
| Topic 6 | close | school | sir | march | till | shut | delhi | school_colleg | govt | govern |
| Topic 7 | peopl | hashtagcorona | world | time | spread | hashtagcovid | india | infect | think | come |
| Topic 8 | hashtagcovid | hashtagcorona | prevent | stay | safe | avoid | spread | way | precaut | india |
| Topic 9 | hashtagcorona | hashtagcovid | cover | true_worship_cure_incur | india | work | new | kind | patient | report |
| Topic 10 | market | world | india | hashtagcorona | economi | hashtagcovid | time | impact | global | peopl |
| Topic 11 | cancel | travel | book | flight | ticket | india | outbreak | march | refund | mentionindigo |
| Topic 12 | test | posit | hashtagcovid | wife | quarantin | hashtagcorona | employe | negat | come | hashtagcovid_ |
| Topic 13 | hai | se | hashtagcorona | bhi | ke | ka | ki | sir | ho | hi |
| Topic 14 | spread | work | india | peopl | sir | health | travel | home | hashtagcovid | world |
| Topic 15 | test | india | patient | hospit | indian | hashtagcovid | case | quarantin | peopl | posit |
| India (Period 2): 16-18 January 2021 | | | | | | | | | |
| Topic 1 | test | sir | travel | school | delhi | student | need | class | month | posit |
| Topic 2 | dose | vaccin | bharat_biotech | crore | avail | serum_institut | rbi_extend_moratorium_end | peopl | pm | import |
| Topic 3 | vaccin | india | pm | today | thank | warrior | scientist | world | fight | start |
| Topic 4 | test | hashtagcovid | posit | lab | new | time | pandem | ice_cream | symptom | player |
| Topic 5 | vaccin | die | norway | pfizer | receiv | peopl | administ | phase_free_cost | centr_bea | hashtagcovid |
| Topic 6 | vaccin | drive | india | begin | world | biggest | hashtagcovid | largest | phase | today |
| Topic 7 | vaccin | india | peopl | hashtagcovid | time | world | pandem | start | year | countri |
| Topic 8 | vaccin | state | drive | maharashtra | hashtagcovid | death | till | depart | januari | today |
| Topic 9 | vaccin | new | india | hashtagcovid | combat_clap_beat_plat | light_lamp_c | pandem | post | impact | case |
| Topic 10 | vaccin | delhi | worker | receiv | health | drive | india | aiiim | hashtagcovid | report |
| Topic 11 | vaccin | india | pm | launch | drive | modi | rollout | pan | world | today |
| Topic 12 | vaccin | hashtagcovid | patient | peopl | hospit | delhi | caller_tune | studi | rapid_blood_test | pandem |
| Topic 13 | hai | vaccin | india | pulwama | happen | se | bhi | ki | ho | modi |
| Topic 14 | vaccin | hospit | worker | health | dr | receiv | drive | launch | doctor | hashtagcovid |
| Topic 15 | case | new | india | hashtagcovid | death | report | total | posit | hour | activ |
| India (Period 3): 12-14 April 2021 | | | | | | | | | |
| Topic 1 | state | peopl | india | vaccin | govt | case | hashtagcovid | situat | govern | hospit |
| Topic 2 | hai | se | exam | bhi | ki | ke | kya | nahi | ko | aur |
| Topic 3 | exam | student | board | cancel | case | postpon | class | cbse | sir | situat |
| Topic 4 | ralli | elect | peopl | india | spread | polit | case | mentionamitshah | pm | follow |
| Topic 5 | vaccin | india | hashtagcovid | dose | sputnik | approv | use | countri | emerg | govt |
| Topic 6 | case | india | new | hashtagcovid | death | report | record | hour | posit | lakh |
| Topic 7 | kumbh | peopl | spread | kumbh_mela | case | india | gather | test | haridwar | hashtagcovid |
| Topic 8 | vaccin | peopl | hashtagcovid | test | time | posit | infect | patient | home | case |
| Topic 9 | vaccin | hashtagcovid | time | peopl | pandem | india | help | work | sir | year |
| Topic 10 | case | hashtagcovid | lockdown | india | maharashtra | wave | state | surg | impos | rise |
| Topic 11 | hospit | patient | bed | help | need | hashtagcovid | urgent | posit | contact | delhi |
| Topic 12 | case | state | elect | kumbh | delhi | meet | hashtagcovid | situat | india | uttarakhand |
| Topic 13 | vaccin | hashtagcovid | follow | mask | peopl | polic | proper | wear_mask | case | hospit |
| Topic 14 | india | peopl | case | april | daili | date | attend_million_godi | attend_godi_media_label | media_label_devot | jihad_akumbh_mela |
| Topic 15 | test | posit | hashtagcovid | report | negat | case | hospit | home | peopl | time |